\documentclass[twocolumn]{article}

\usepackage{PRIMEarxiv}

\usepackage[utf8]{inputenc}
\usepackage[T1]{fontenc}
\usepackage{hyperref}
\hypersetup{
  colorlinks=true,
  linkcolor=blue,
  citecolor=blue,
  urlcolor=blue
}
\usepackage{url}
\usepackage{booktabs}
\usepackage{amsfonts}
\usepackage{amsmath}
\usepackage{amssymb}
\usepackage{mathtools}
\usepackage{amsthm}
\usepackage{microtype}
\usepackage{graphicx}
\usepackage{subcaption}
\usepackage{fancyhdr}
\usepackage{xcolor}
\usepackage{listings}
\usepackage{algorithm}
\usepackage{algorithmic}
\usepackage{xspace}
\usepackage{pifont}

\usepackage[capitalize,noabbrev]{cleveref}

\usepackage{amsmath,amsfonts,bm}

\def\eqref#1{equation~\ref{#1}}

\def\1{\bm{1}}

\DeclareMathAlphabet{\mathsfit}{\encodingdefault}{\sfdefault}{m}{sl}
\SetMathAlphabet{\mathsfit}{bold}{\encodingdefault}{\sfdefault}{bx}{n}

\newcommand{\corpgen}{\textsc{CorpGen}{}\xspace}
\newcommand{\react}{\textsc{ReAct}{}\xspace}

\theoremstyle{plain}

\theoremstyle{definition}

\theoremstyle{remark}

\title{\corpgen: Simulating Corporate Environments with Autonomous Digital Employees in Multi-Horizon Task Environments}

\author{
  \textbf{Abubakarr Jaye}$^{\dagger}$, \textbf{Nigel Boachie Kumankumah}, \textbf{Chidera Biringa}\\[0.3em]
  Anjel Shaileshbhai Patel, Sulaiman Vesal, Dayquan Julienne, Charlotte Siska, \\
  Manuel Ra\'{u}l Mel\'{e}ndez Luj\'{a}n, Anthony Twum-Barimah, Mauricio Velazco, Tianwei Chen \\[0.3em]
  Microsoft Corporation \\[0.2em]
  {\footnotesize $^{\dagger}$Project Lead, Correspondence: ajaye@microsoft.com}
}

\begin{document}

\twocolumn[
  \maketitle
  \vspace{1.5em}
]

\begin{abstract}
Long-horizon reasoning is a key challenge for autonomous agents, yet existing benchmarks evaluate agents on single tasks in isolation. Real organizational work requires managing many concurrent long-horizon tasks with interleaving, dependencies, and reprioritization. We introduce Multi-Horizon Task Environments (MHTEs): a distinct problem class requiring coherent execution across dozens of interleaved tasks (45+, 500--1500+ steps) within persistent execution contexts spanning hours. We identify four failure modes that cause baseline CUAs to degrade from 16.7\% to 8.7\% completion as load scales 25\%$\to$100\%, a pattern consistent across three independent implementations. These failure modes are context saturation ($O(N)$ vs $O(1)$ growth), memory interference, dependency complexity (DAGs vs.\ chains), and reprioritization overhead.

We present \corpgen, an architecture-agnostic framework addressing these failures via hierarchical planning for multi-horizon goal alignment, sub-agent isolation preventing cross-task contamination, tiered memory (working, structured, semantic), and adaptive summarization. \corpgen simulates corporate environments through digital employees with persistent identities and realistic schedules. Across three CUA backends (UFO2, OpenAI CUA, hierarchical) on OSWorld Office, \corpgen achieves up to 3.5$\times$ improvement over baselines (15.2\% vs 4.3\%) with stable performance under increasing load, confirming that gains stem from architectural mechanisms rather than specific CUA implementations. Ablation studies show experiential learning provides the largest gains.

\end{abstract}
\vspace{0.5em}

\begin{figure}[t]
\centering
\includegraphics[width=\columnwidth]{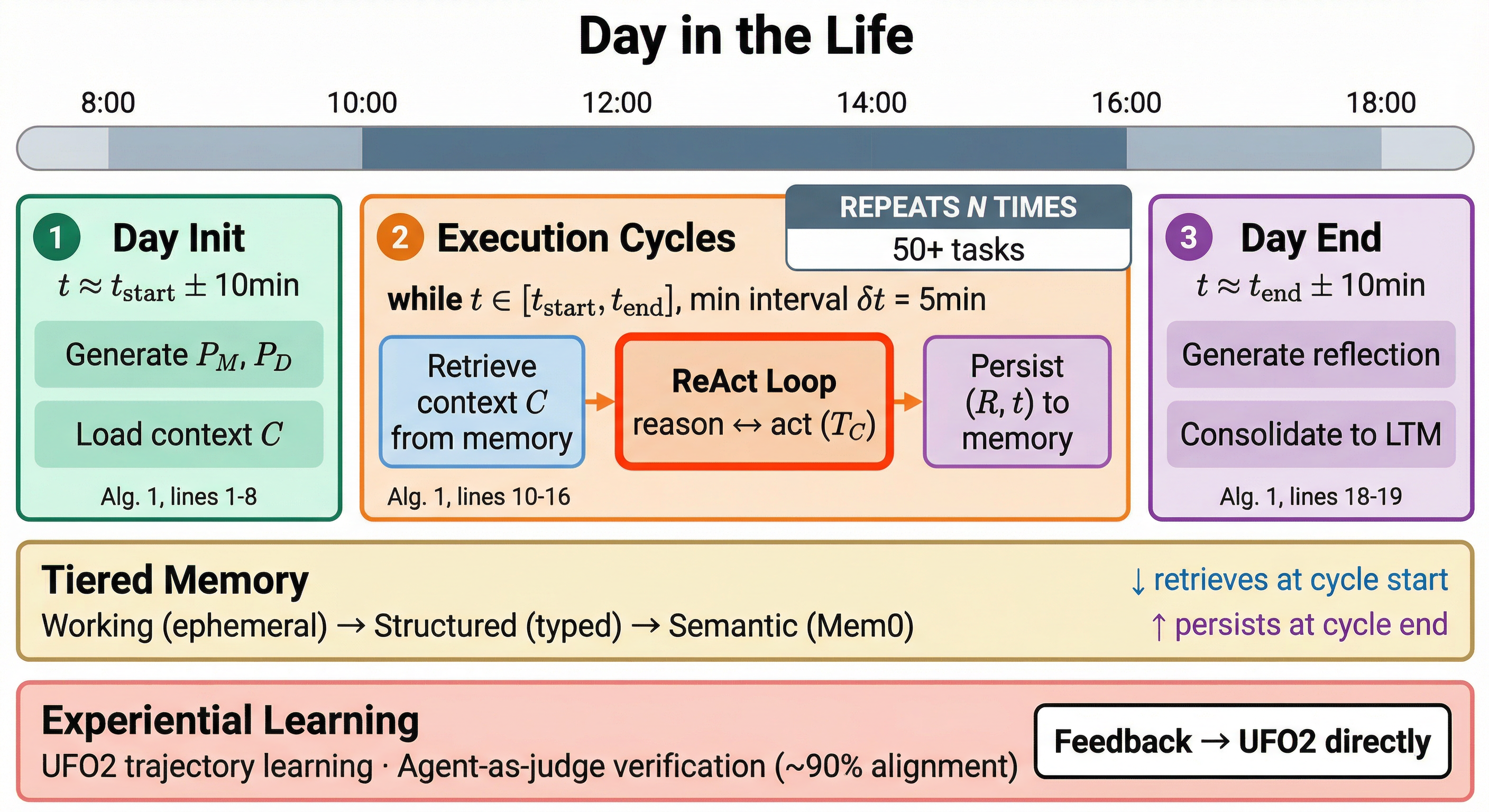}
\caption{A Digital Employee's Workday. The agent begins with Day Init, loading memory and generating a daily plan. It then runs repeated Execution Cycles, where it retrieves context, reasons and acts, and persists results while completing many interleaved tasks. At Day End, the agent reflects on its actions and consolidates experience into long-term memory, enabling coherent operation over hours despite context resets.}
\label{fig:day_in_the_life}
\end{figure}

\section{Introduction}
\label{introduction}

Long-horizon reasoning is widely regarded as a key challenge for autonomous agents, with benchmarks like ALFWorld~\cite{shridharalfworld} and WebArena~\cite{zhouwebarena} testing agents' ability to complete extended action sequences requiring 10--20 steps. However, these benchmarks share a critical limitation: they evaluate agents on \textit{one task at a time}. Real organizational work requires managing \textit{many} long-horizon tasks concurrently, with interleaved execution, cross-task dependencies, and continuous reprioritization. This distinction is not merely quantitative; it represents a fundamentally different problem class with distinct failure modes.

We introduce \textbf{Multi-Horizon Task Environments (MHTEs)}: environments containing dozens of concurrent tasks, each requiring its own long-horizon plan (10--30+ steps per task), evaluated within a single persistent execution context spanning hours. Unlike single-task long-horizon benchmarks that test \textit{sequential persistence} (can the agent complete one extended task?), MHTEs test \textit{concurrent persistence} (can the agent maintain coherent execution across many interleaved tasks simultaneously?). We demonstrate empirically that single-task long-horizon agents experience catastrophic performance degradation in MHTEs: baseline computer using agents (CUAs) degrade from 16.7\% task completion at 25\% load to 8.7\% at 100\% load, a pattern consistent across three independent CUA implementations. This degradation stems from four fundamental architectural mismatches: bounded context window saturation (context grows $O(N)$ with task count rather than $O(1)$), cross-task memory interference (information from one task contaminates reasoning about others), dependency graph complexity (DAGs versus linear chains), and reprioritization overhead ($O(N)$ versus $O(1)$ per-cycle decision complexity).

To operate within MHTEs, we define \textbf{MOMA (Multi-Objective Multi-Horizon Agent)} capabilities: agents must (1) simultaneously manage multiple objectives, (2) plan over multiple long horizons, (3) track state across many concurrent task threads, and (4) dynamically reprioritize based on progress and emerging constraints. We operationalize these capabilities through autonomous \textbf{digital employees}: LLM-powered agents instantiated with persistent identities, role-specific expertise, and realistic work schedules. These agents interact with enterprise applications through Graphical User Interface (GUI) automation and operate within MHTEs, maintaining coherence over hours of execution across 45+ concatenated tasks. Agent identity and memory persist across daily boundaries.

\begin{table}[t]
\caption{Why MHTEs Are Qualitatively Different. MHTEs introduce fundamental architectural challenges beyond single-task long-horizon benchmarks, leading to distinct failure modes.}
\label{tab:mhte_comparison}
\centering
\small
\begin{tabular}{@{}lcc@{}}
\toprule
\textbf{Dimension} & \textbf{Long-Horizon} & \textbf{MHTE} \\
\midrule
Number of tasks & 1 & 45+ \\
Steps per task & 10--20 & 10--30+ \\
Total steps & 10--20 & 500--1500+ \\
Concurrency & None & High \\
Cross-task dependencies & N/A & Common \\
Reprioritization & Rare & Frequent \\
\midrule
\multicolumn{3}{c}{\textit{Architectural Challenges}} \\
\midrule
Context growth & $O(1)$ & $O(N)$ \\
Memory interference & None & High \\
Decision complexity & $O(1)$ & $O(N)$ \\
Dependency structure & Linear/Tree & DAG/Cyclic \\
\midrule
\multicolumn{3}{c}{\textit{Failure Modes}} \\
\midrule
Context saturation & Rare & Common \\
Task conflation & None & Frequent \\
Priority thrashing & None & Common \\
Baseline degradation & Stable & Catastrophic \\
\bottomrule
\end{tabular}
\end{table}

Sustaining execution at this scale introduces challenges of goal alignment across thousands of reasoning steps, context degradation as token windows reset, state preservation across execution boundaries, and cross-task dependency management. To address these, we present \corpgen, an architecture-agnostic framework that realizes MOMA capabilities through targeted architectural mechanisms. Our contributions are threefold:

\begin{itemize}
    \item We introduce Multi-Horizon Task Environments (MHTEs) as a distinct problem class from single-task long-horizon reasoning, identify four fundamental failure modes (context saturation, memory interference, dependency complexity, reprioritization overhead), and demonstrate empirically that baseline CUAs degrade catastrophically as task load increases (16.7\%$\rightarrow$8.7\% completion).

    \item We define Multi-Objective Multi-Horizon Agent (MOMA) capabilities required for MHTEs and present \corpgen, an architecture-agnostic framework that addresses each failure mode through hierarchical planning, sub-agent isolation, tiered memory, and adaptive summarization. Evaluation across three CUA backends demonstrates consistent improvements (up to 3.5$\times$ baseline), validating that gains stem from architectural mechanisms rather than specific CUA implementations.

    \item We show through ablation studies that experiential learning provides the largest performance gains among architectural components, and observe in preliminary evaluation experiments that artifact-based judgment achieves substantially higher agreement with human labels than trace-based LLM judgment, suggesting that standard evaluation methods may underestimate long-horizon agent performance.
\end{itemize}

\section{Related Work}
\label{related-work}

\textbf{LLM-Powered Agents.} Park et al.'s Generative Agents~\cite{park2023generative} introduced memory, planning, and reflection mechanisms. The ReAct framework~\cite{yao2022react} combines reasoning and acting in iterative processes. Reflexion~\cite{shinn2023reflexion} implements verbal reinforcement learning for agent improvement. Laban et al. \cite{laban2025llms} identified challenges in multi-turn conversations, highlighting the importance of context management. \corpgen builds on these approaches with specialized architectures for corporate simulation. Existing frameworks such as UFO2~\cite{zhang2025ufo2} provide strong GUI-level task execution but lack the higher-level orchestration needed for sustained multi-task, multi-horizon operation.

\textbf{Memory Systems.}
Li et al. introduced MemOS \cite{li2025memos} with metadata-enriched memory objects. Mem0 \cite{chhikara-2025} offers lightweight, production-ready solutions. Wang and Chen's MIRIX \cite{wang-2025} implements multi-agent memory with six distinct types. Xu et al.'s A-MEM \cite{xu2025amem} proposes adaptive, self-organizing memory inspired by Zettelkasten. \corpgen incorporates structured memory management with specialized types for corporate knowledge.

\textbf{Multi-Agent Collaboration.} Hong et al.'s MetaGPT \cite{hong2023metagpt} encodes Standardized Operating Procedures into multi-agent workflows. Qian et al.'s ChatDev \cite{qian2024chatdev} demonstrates virtual software companies with role-based agents. Pan et al.'s AgentScope \cite{pan2024very} provides actor-based distributed architecture for large-scale simulations. Yang et al. \cite{Yang2024LLMbasedMS} discuss organizational structures in multi-agent systems. \corpgen differs by enabling emergent collaboration through independent digital employees interacting via standard communication channels rather than predefined coordination protocols.

\textbf{Long-Horizon Tasks vs. Multi-Horizon Task Environments.} Prior work describes ``long-horizon tasks'' as single tasks with extended action sequences. We study a fundamentally different regime: long \textit{sets} of sequences, pursued concurrently and interactively. Classical long-horizon benchmarks (e.g., ALFWorld~\cite{shridharalfworld}, WebArena~\cite{zhouwebarena}) test \textit{sequential persistence}: whether an agent can complete one complex task through many steps. MHTEs test \textit{concurrent persistence}: whether agents can maintain coherent execution across dozens of interleaved tasks simultaneously (Table~\ref{tab:mhte_comparison}). We identify four failure modes driving this gap: context saturation, memory interference, dependency complexity, and reprioritization overhead. We validate them empirically in Section~\ref{results}.

\textbf{Environment Interaction and Computer Using Agents.}
For agents to realistically simulate employees in a corporate setting, they must interact with digital environments in human-like ways. Recent work by OpenAI describes computer using agents (CUAs) \cite{openai-no-date} that can control computers through the GUI, performing actions like clicking, typing, and navigating interfaces. These systems face challenges in UI understanding and error handling. \corpgen provides agents with a controlled digital workplace including feedback loops for action observation, error handling, and human-like behaviors (realistic typing speeds, work schedules).

Prior work on LLM-powered agents, memory systems, and multi-agent collaboration has made substantial progress on isolated long-horizon tasks, predefined workflows, or structured coordination protocols. However, these settings do not evaluate sustained execution across dozens of concurrent, interdependent long-horizon tasks within a single persistent context. In contrast, our work studies a distinct regime in which agents must manage multiple objectives simultaneously, track cross-task state over extended execution, and dynamically reprioritize under bounded context windows, motivating the need for Multi-Horizon Task Environments, MOMA capabilities, and evaluation methods that account for sustained, concurrent execution.

\section{Methodology}
\label{problem-formulation}

\subsection{Problem Setting}

A Multi-Horizon Task Environment (MHTE) consists of a set of tasks $T = \{T_1, \ldots, T_N\}$, where each task $T_i$ requires a sequence of 10--30+ dependent actions. In an MHTE, multiple tasks are active concurrently, tasks may share stateful resources such as files, applications, and communication channels, and progress on one task can create or resolve constraints for others. Agents are evaluated on aggregate task completion within a single persistent execution context spanning hours of execution.

Unlike multi-task benchmarks that sample independent tasks, MHTEs require sustained cross-task state tracking and dynamic reprioritization under a bounded context window. Classical long-horizon benchmarks (e.g., ALFWorld~\cite{shridharalfworld}, WebArena~\cite{zhouwebarena}) evaluate agents on one extended task at a time, testing sequential persistence. MHTEs test \textit{concurrent persistence}: the ability to maintain coherent execution across many interleaved long-horizon tasks simultaneously.

\subsection{Why Single-Task Long-Horizon Agents Fail in MHTEs}

While single-task long-horizon agents can successfully complete extended action sequences when evaluated in isolation, they experience catastrophic performance degradation in MHTEs. Our results (Section~\ref{results}, Table~\ref{tab:load_scalability}) demonstrate this empirically: baseline computer using agents (CUAs) degrade from 16.7\% task completion at 25\% load to 8.7\% at 100\% load. This degradation is not merely a capacity problem; it reflects fundamental architectural mismatches between single-task reasoning and multi-task execution. We identify four key failure modes:

\noindent\textbf{1. Bounded Context Window Saturation.}
In single-task settings, context accumulates linearly with task progress: observations, actions, and intermediate results stack sequentially within the context window. In MHTEs, context requirements grow multiplicatively. Consider two concurrent tasks, each requiring 500 tokens of state. Sequential execution needs 500 tokens at a time; concurrent execution requires maintaining both task states simultaneously (1000 tokens) plus task-switching overhead (routing decisions, context restoration, cross-task dependency tracking). As $N$ tasks accumulate, the context window must hold $O(N)$ active task states rather than $O(1)$, rapidly exceeding capacity and forcing lossy summarization or state eviction.

\noindent\textbf{2. Cross-Task Memory Interference.}
When multiple tasks share a context window, information from one task contaminates reasoning about others. For example, an agent working on both a sales report and an engineering report may conflate data between them (``update the Q4 spreadsheet'': which one?). Single-task agents assume context is task-specific; MHTE agents must explicitly partition and retrieve task-scoped memory. Without isolation mechanisms, retrieval becomes ambiguous and actions misfire. This interference compounds over time: each context-polluted decision creates incorrect state that propagates to future cycles.

\noindent\textbf{3. Dependency Graph Complexity.}
Single-task long-horizon problems have linear or tree-structured dependency graphs: complete step $A$, then step $B$, then step $C$. MHTEs introduce cross-task dependencies that form directed acyclic graphs (DAGs) or even cyclic constraints. For instance, Task A (write report) depends on Task B (analyze data), which depends on Task C (request data from colleague), which may depend on Task A being drafted first to clarify requirements. Agents must perform topological reasoning over the dependency graph to determine valid execution orderings, maintain partial completion states for blocked tasks, and dynamically re-evaluate priorities as dependencies resolve. Single-task planners lack this multi-objective scheduling capability.

\noindent\textbf{4. Reprioritization Overhead.}
In single-task execution, the next action is determined by the current task state alone: ``What is the next step toward completing this task?'' In MHTEs, every action requires re-evaluating priorities across all active tasks: ``Which task should I work on next, given current progress, deadlines, blockers, and dependencies?'' This introduces $O(N)$ decision complexity at each cycle versus $O(1)$ for single tasks. Without hierarchical planning to amortize these decisions, agents waste reasoning capacity on task selection rather than task execution, leading to thrashing and stalled progress.

\noindent\textbf{Empirical Evidence.}
Table~\ref{tab:load_scalability} demonstrates these failure modes quantitatively. As task load increases from 25\% (12 tasks) to 100\% (46 tasks), baseline CUAs show consistent degradation across architectures (UFO2: 8.3\%$\rightarrow$4.3\%; computer-use-preview: 16.7\%$\rightarrow$8.7\%; hierarchical: 25.0\%$\rightarrow$14.1\%). This pattern holds across three independent CUA implementations, indicating that the degradation stems from the problem structure (concurrent multi-task execution) rather than implementation details. In contrast, \corpgen maintains or improves performance at higher loads by directly addressing each failure mode through its architecture (Section~\ref{architecture}).

\subsection{Required Capabilities: Multi-Objective Multi-Horizon Agents (MOMA)}

Operating effectively in MHTEs requires a qualitatively different set of agent capabilities beyond single-task long-horizon reasoning. We define \textbf{Multi-Objective Multi-Horizon Agent (MOMA)} capabilities as:

\begin{enumerate}
    \item \textbf{Simultaneous Objective Management:} Maintain coherent progress toward multiple goals concurrently, with explicit partitioning of task-specific state to prevent memory interference.

    \item \textbf{Multi-Horizon Planning:} Decompose strategic objectives (weeks-months) into tactical plans (days) and operational actions (per-cycle), with dynamic re-planning as task states evolve.

    \item \textbf{Cross-Task State Tracking:} Persist critical information across task boundaries and context resets, using tiered memory to balance working context, structured knowledge, and semantic retrieval.

    \item \textbf{Dynamic Reprioritization:} Evaluate task priorities based on progress, dependencies, deadlines, and emerging constraints, scheduling execution to maximize aggregate completion.
\end{enumerate}

These capabilities directly address the four failure modes identified above: tiered memory prevents context saturation, isolated sub-agents prevent memory interference, hierarchical planning manages dependency graphs, and structured prioritization reduces reprioritization overhead. Section~\ref{architecture} presents \corpgen's architecture for realizing these capabilities.

\subsection{\corpgen~Framework}
\label{architecture}

Section~\ref{problem-formulation} identified four fundamental failure modes that cause single-task long-horizon agents to degrade catastrophically in Multi-Horizon Task Environments (MHTEs): bounded context window saturation, cross-task memory interference, dependency graph complexity, and reprioritization overhead. This section presents \corpgen, an architecture-agnostic framework that addresses each failure mode through targeted mechanisms while remaining modular and CUA-independent.

\corpgen realizes Multi-Objective Multi-Horizon Agent (MOMA) capabilities through four core architectural mechanisms, each addressing a specific failure mode: (1) \textbf{hierarchical planning} decomposes strategic objectives into tactical plans and operational actions, reducing reprioritization overhead from $O(N)$ to amortized $O(1)$ and managing dependency graphs through explicit milestone tracking; (2) \textbf{sub-agents as tools} isolate complex operations (research, GUI automation) in dedicated contexts, preventing cross-task memory interference by returning only structured results to the host agent; (3) \textbf{tiered memory} persists critical state across working memory (intra-cycle), structured long-term memory (plans, summaries), and semantic memory (embeddings), addressing context saturation by enabling selective retrieval rather than full context retention; and (4) \textbf{adaptive summarization} actively manages context growth through rule-based compression, preserving critical information while bounding token consumption.

These core mechanisms are complemented by cognitive tools for structured reasoning, experiential learning through trajectory-based feedback, and support for emergent multi-agent collaboration. Together, this architecture enables reliable MHTE operation while remaining modular: individual components can be improved or replaced (e.g., swapping CUA implementations) without system-wide redesign.

\begin{figure*}[t]
\centering
\includegraphics[width=\textwidth]{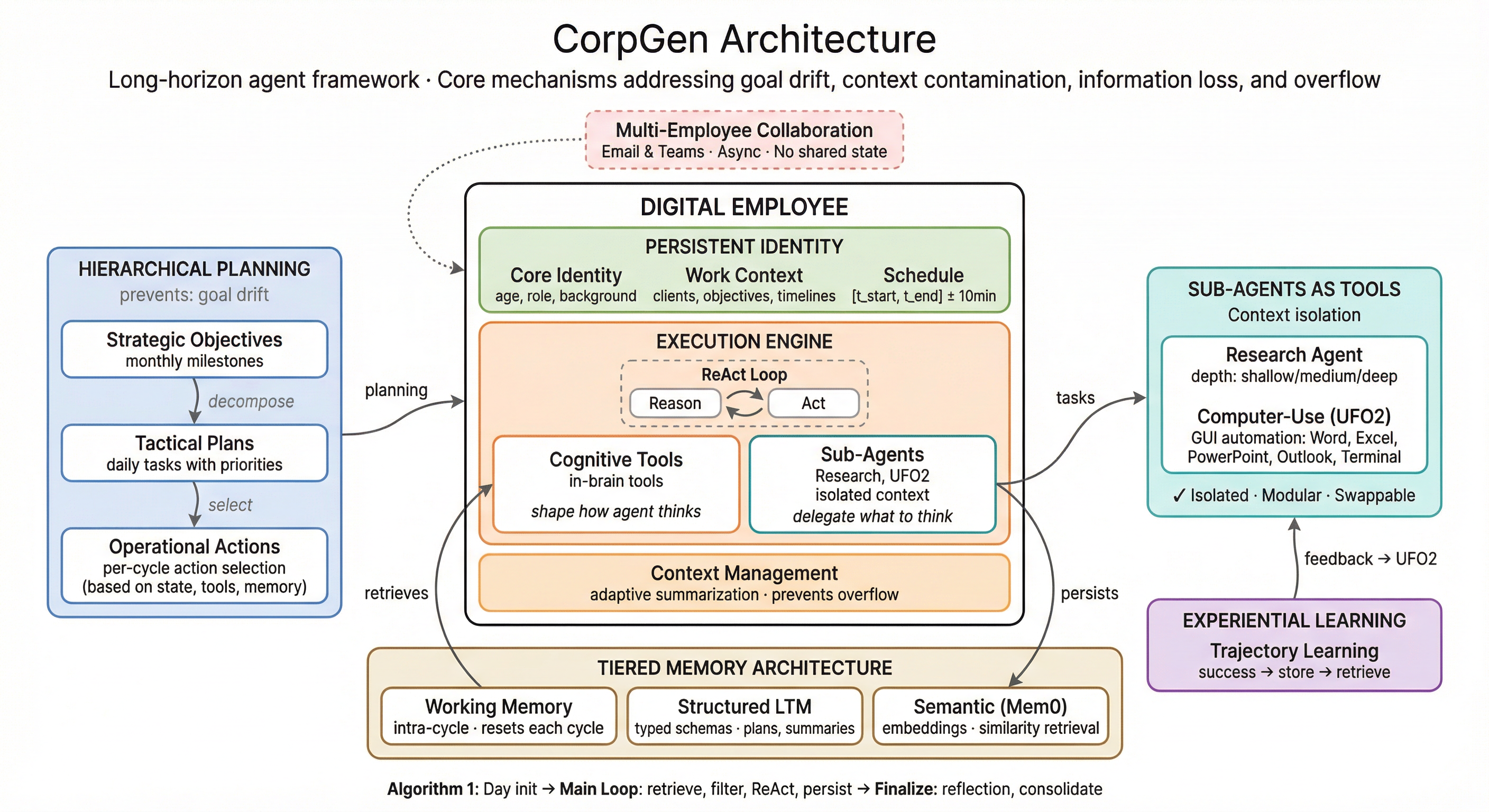}
\caption{\corpgen Architecture. Four core mechanisms address long-horizon challenges: \textbf{Hierarchical Planning} (\S\ref{subsec:hierarchical_planning}) decomposes strategic objectives (monthly) into tactical plans (daily) and operational actions (per-cycle) to prevent goal drift. The \textbf{Execution Engine} runs a \react loop (Algorithm~\ref{alg:agent_execution}, lines 11--15) invoking \textbf{Cognitive Tools} (\S\ref{subsec:cognitive_tools}) within host context and \textbf{Sub-Agents as Tools} (\S\ref{subsec:subagents}) in isolated context scopes. \textbf{Context Management} (\S\ref{subsec:context_management}) prevents overflow via adaptive summarization. \textbf{Tiered Memory} (\S\ref{subsec:memory_arch}) prevents information loss across working memory (intra-cycle), structured LTM (plans, summaries), and semantic memory (Mem0 embeddings). \textbf{Experiential Learning} (\S\ref{subsec:experiential_learning}) captures successful execution patterns via UFO2 trajectory learning, with feedback routed directly to UFO2's app agent. External interfaces include enterprise applications via UFO2 GUI and communication channels enabling \textbf{Emergent Collaboration} (\S\ref{subsec:collaboration}).}
\label{fig:architecture}
\end{figure*}

\subsubsection{Hierarchical Planning for Long-Term Consistency}
\label{subsec:hierarchical_planning}

\corpgen maintains strategic coherence through hierarchical goal decomposition at three temporal scales:

\noindent\textbf{Strategic Objectives (Monthly).} High-level goals with milestone dates derived from agent identity and role responsibilities. For example, a Principal Researcher's monthly objective might specify ``GPT-5 preliminary research report'' with concrete deliverables: research findings by week 2, draft report by week 3, final presentation by month end.

\noindent\textbf{Tactical Plans (Daily).} Each day, strategic objectives decompose into 6-12 actionable tasks specifying target applications (Word, Excel, PowerPoint, Outlook, Edge), priority rankings, and dependencies. Daily plans align with monthly milestones while adapting to immediate context.

\noindent\textbf{Operational Actions (Per-Cycle).} Each execution cycle, the agent selects actions from daily plans based on current state, available tools, and retrieved memory context. Cycles run to completion regardless of duration.

This hierarchical structure ensures that low-level decisions remain grounded in high-level intent. When an agent completes a task or encounters blockers, plan update operations propagate changes through the hierarchy: task completion triggers daily progress updates and milestone tracking, while blockers cause priority adjustments and potential escalation to monthly plan revision. New information from research or communication feeds back into planning, enabling responsive adaptation while preserving strategic alignment.

Hierarchical planning in \corpgen is updated within the execution loop (Algorithm~\ref{alg:agent_execution}). A high-level plan is generated at initialization and decomposed into daily and per-cycle tasks. Plan updates occur when the agent detects that the current plan is no longer consistent with the observed environment state. These events trigger updates to the current daily plan, which may in turn revise higher-level objectives when progress deviates from expectations. Updated plans propagate downward by refreshing the active task queue used in subsequent execution cycles. This event-driven update scheme enables adaptation to execution outcomes without continuous replanning.

\subsubsection{Sub-Agents as Tools for Context Isolation}
\label{subsec:subagents}

\corpgen isolates complex operations through modular sub-agents wrapped as tools, preventing cross-task context contamination. Sub-agents are autonomous reasoning agents that execute complex, multi-step operations in isolated contexts, returning only structured results to the host agent. This differs from cognitive tools (Section~\ref{subsec:cognitive_tools}), which are functions that force the model to think in a specific way by requiring structured outputs; they shape reasoning patterns rather than delegating to independent agents.

\noindent\textbf{Research Agent.} A dedicated reasoning agent for depth-configurable research (shallow/medium/deep corresponding to 1/2/3 iteration steps). The research agent navigates web browsers, collects information, and synthesizes findings into structured reports. Critically, it operates within an isolated context scope: the host agent provides a research query and receives a structured report, without the research agent's intermediate reasoning polluting the host's context window.

\noindent\textbf{Computer Using Agent (UFO2).} UFO2 \cite{zhang2025ufo2} is the primary GUI automation agent, providing vision-based control for Windows applications via accessibility APIs. The sub-agent architecture is modular: the computer using agent (CUA) can be replaced with alternative CUAs \cite{gonzalez-pumariega-2025-scaling}, enabling updates without changing host agent architecture.

\subsubsection{Tiered Memory Architecture}
\label{subsec:memory_arch}
\corpgen implements a tiered memory system with three layers: \textbf{Working Memory (Intra-Cycle)}, which holds active task state and immediate reasoning and resets each cycle; \textbf{Structured Long-Term Memory}, which stores typed artifacts such as plans, action summaries, and reflections; and \textbf{Semantic Memory}, which supports similarity-based retrieval over unstructured past context using Mem0 \cite{chhikara-2025}. At the start of each execution cycle, memory retrieval follows a fixed priority ordering, favoring recently accessed entries, task-critical records marked as important (e.g., task state changes, plan updates, and failures), and semantically relevant past context. A small, configurable set of the top-ranked entries is injected into working memory for the cycle, ensuring that critical information is accessible while keeping the context bounded.

\subsubsection{Summarization and Context-Management Modules}
\label{subsec:context_management}

To bound long execution contexts, \corpgen applies rule-based adaptive summarization within the execution loop shown in Algorithm~\ref{alg:agent_execution}. Context is divided into critical content, including tool invocations, task state changes, plan updates, and error or recovery signals, and routine content, including intermediate observations and transient reasoning. When context length exceeds a fixed number of tokens (4k), critical content is preserved verbatim while routine content is compressed into a short structured summary capturing decisions, blockers, and current application state. If context growth continues, routine content is further compressed while all task-level state is retained. Summaries are written to memory and retrieved at the start of subsequent execution cycles, with rare over-compression failures handled by the retry and skip policy.

\textbf{Execution Cycle with Context Management.}
Digital employees operate on time-bounded schedules (default 8--18h with a $\pm 10$-minute variance) and execute work in discrete cycles. Each cycle retrieves context and memory, selects the next task, and completes a full \react loop while persisting reflections; cycles always run to completion because sub-agent invocations (e.g., UFO2 GUI automation) may take minutes to hours, with a minimum five-minute interval between cycle starts. Task interleaving occurs across cycles rather than within them by updating and reprioritizing the task queue after each cycle, and execution follows the \react paradigm \cite{yao2022react}, alternating between reasoning and tool-based actions while isolating sub-agent state from the main reasoning process..

\begin{algorithm}
\small
\caption{Digital Employee Execution Loop}
\label{alg:agent_execution}
\begin{algorithmic}[1]
\REQUIRE Agent identity $I$, Working hours $[t_{start}, t_{end}]$, Interval $\delta t$
\ENSURE Completed workday with persistent memory updates

\STATE \textbf{// Day Initialization}
\STATE Apply $\pm 10$ min variance to $t_{start}, t_{end}$ for behavioral realism
\IF{no monthly plan exists for current month}
    \STATE Generate monthly plan $P_M$ with strategic objectives and milestones
    \STATE Store $P_M$ in structured long-term memory
\ENDIF
\STATE Generate daily plan $P_D$ from monthly objectives and current context
\STATE Store $P_D$ in structured long-term memory

\STATE \textbf{// Main Execution Loop}
\WHILE{current time $t \in [t_{start}, t_{end}]$}
    \STATE Retrieve context $C$ from stratified memory (plans, past actions, semantic)
    \STATE Filter tools $T_C$ based on context $C$ and current goals
    \STATE Execute \react loop: reason about state, invoke tools from $T_C$, produce result $R$
    \STATE \textit{(Summarization triggers automatically if context approaches token limit)}
    \STATE Store action summary $(R, t)$ in memory with metadata
    \STATE Wait for minimum interval $\delta t$ before next cycle
\ENDWHILE
\STATE \textbf{// Day Finalization}
\STATE Generate end-of-day reflection summarizing accomplishments
\STATE Consolidate experiences into long-term memory structures
\end{algorithmic}
\end{algorithm}

\textbf{Retry and Skip Policy for Resilience.} Long-horizon execution must handle infeasible tasks and persistent failures. \corpgen uses a retry-and-skip policy to maintain progress: tasks that fail after three attempts, each capped at 30 iterations, are skipped rather than blocking execution. This prevents the agent from stalling on infeasible or transiently failing tasks, which is critical for long-horizon evaluation where a single stalled task could halt all subsequent execution. Skipped tasks are logged for analysis and counted as failures, allowing the system to continue processing the remaining queue without inflating the success rate.

\textbf{Employee Identity as Stable Context.} Digital employees maintain consistent behavior through identity encodings that persist across execution cycles. These encodings capture core personal and professional attributes, work context such as objectives and client relationships, and tool assignments with time-bounded daily routines. This identity layer remains constant across context resets, ensuring behavioral consistency even as working memory contents change.

\subsection{Cognitive Tools for Structured Operations}
\label{subsec:cognitive_tools}
While sub-agents (Section~\ref{subsec:subagents}) are autonomous reasoning agents with isolated contexts, \textbf{cognitive tools} shape the agent’s reasoning within a single context by enforcing structured thought patterns. Rather than performing independent reasoning, these tools constrain how the agent reasons by requiring explicit structure in planning, progress tracking, and reflection. The agent decides when to invoke these tools based on its state and goals, with system guidance encouraging their use at appropriate points in the execution cycle.

We use three classes of cognitive tools. \textbf{Planning tools} require the agent to explicitly decompose objectives into structured plans and task lists, supporting consistent goal formulation and adaptation over time. \textbf{Task tracking tools} force explicit accounting of completed and remaining tasks, enabling systematic progress monitoring across execution cycles. \textbf{Reflection tools} structure end-of-day reasoning around outcomes and lessons learned, consolidating experience into long-term memory. Together, these tools promote disciplined reasoning without introducing additional autonomous agents.

\subsection{Experiential Learning}
\label{subsec:experiential_learning}

Experiential learning in \corpgen uses lightweight reuse of verified demonstrations rather than parameter updates or policy optimization. When a task succeeds, the system distills the execution into a minimal, canonical trajectory capturing context, application state, and correct actions, embeds it using a sentence-transformer model, and indexes it in an application-aware FAISS database. At execution time, the top-k similar demonstrations are retrieved and injected as few-shot examples to bias action selection without changing the planner or model parameters. Retrieval is application-aware: queries are filtered to return only demonstrations from the same application, preventing misleading cross-application retrievals. This approach empirically reduces exploratory actions and accelerates convergence on repeated or structurally similar tasks, while learning remains limited to reusing validated interaction patterns rather than generalizing beyond stored demonstrations.

\begin{figure}[ht]
\centering
\includegraphics[width=\columnwidth]{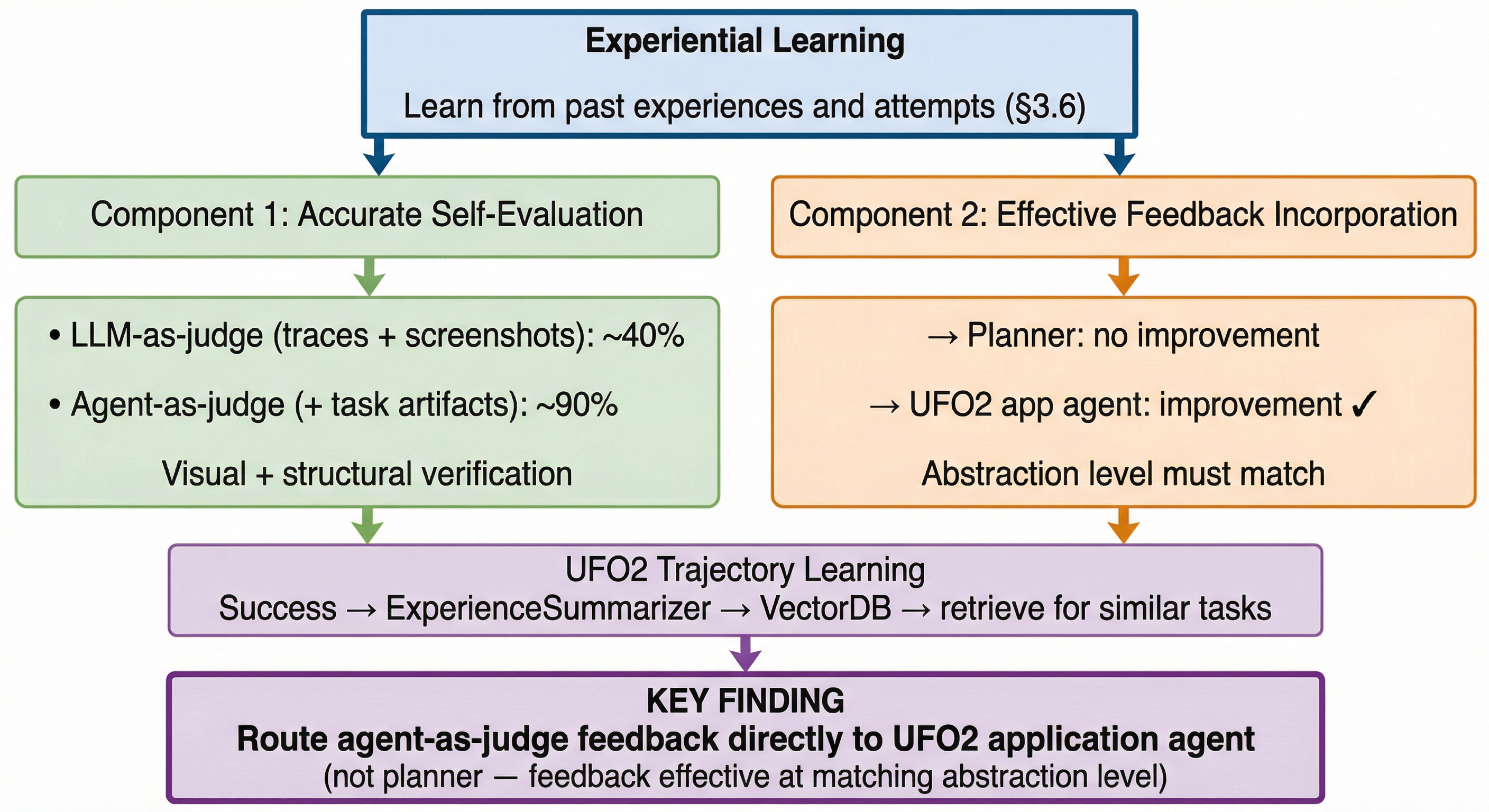}
\caption{Experiential learning cycle. UFO2's trajectory learning captures successful execution patterns and distills them into canonical demonstrations indexed for retrieval. Feedback from successful executions is routed directly to the execution agent, biasing future action selection toward validated patterns.}
\label{fig:experiential_learning}
\end{figure}

\subsection{Emergent Collaboration}
\label{subsec:collaboration}

When multiple digital employees operate in the same environment, collaboration emerges through standard communication channels such as email and Microsoft Teams rather than shared internal state. Messages sent by one employee are retrieved by others during their execution cycles, processed with relevant memory context, and responded to asynchronously, closely mirroring real workplace communication. Coordination adapts under failure: if a communication path breaks, such as an email delivery error, agents reroute messages through alternative channels like Teams to ensure task completion. Over time, these independent interactions give rise to organizational dynamics, including role-based hierarchies, informal collaboration networks, and shared artifacts that function as coordination points. Evaluation of multi-agent collaboration dynamics is left to future work.

\section{Experimental Results}
\label{results}

Section~\ref{problem-formulation} predicted that single-task long-horizon agents would experience catastrophic performance degradation in Multi-Horizon Task Environments (MHTEs) due to four fundamental failure modes. Our evaluation tests three hypotheses: (1) \textbf{Degradation Hypothesis}: Do baseline computer using agents (CUAs) degrade as task load increases, validating the predicted failure modes? (2) \textbf{Architecture Hypothesis}: Does \corpgen's architecture mitigate this degradation by addressing each failure mode? (3) \textbf{Generalization Hypothesis}: Do improvements hold across multiple CUA backends, demonstrating that gains stem from architectural mechanisms rather than specific implementations?

We evaluate these hypotheses through three complementary experiments: task load scalability (Section~\ref{subsec:load_scalability}), component ablation (Section~\ref{subsec:ablation}), and application-specific long-horizon evaluation (Section~\ref{subsec:app_eval}). Results confirm all three hypotheses: baseline CUAs show consistent degradation across implementations, \corpgen maintains or improves performance at higher loads, and gains generalize across three independent CUA backends.

\subsection{Benchmark and Metrics}

To evaluate \corpgen under Multi-Horizon Task Environments (MHTEs), we build on the OSWorld Office benchmark~\cite{xie2024osworld} but adapt it to stress multi-objective, long-horizon behavior. OSWorld was originally designed for atomic or single-task evaluation and is insufficient for testing MOMA capabilities, which require sustained execution across many concurrent tasks. While \corpgen is designed to support persistent execution across multiple simulated workdays, all experiments reported here are run within single-day, high-load sessions for controlled evaluation.

We repurpose OSWorld by concatenating multiple tasks into a single persistent execution context, creating long-running sessions that span hours of interaction. Tasks are grouped in two complementary settings: (1) task load scalability, where the number of concurrent tasks increases from 25\% to 100\%, and (2) per-application long-horizon evaluation, where all tasks within a single application are concatenated into one prompt. This transformation yields a new benchmark tailored specifically to evaluating MHTEs, preserving OSWorld’s task realism while introducing concurrency, dependency tracking, and reprioritization demands absent from the original setup.

Evaluation uses aggregate task completion rate as the primary metric, computed via deterministic artifact comparison against golden reference outputs and supplemented with agent-as-a-judge verification for complex cases. This protocol measures whether agents can sustain coherent execution as task load and horizon length increase, directly testing the architectural goals of \corpgen in realistic multi-task settings.

\noindent\textbf{Experiment Configuration.} All experiments use an Azure OpenAI deployment of \textit{gpt-5.1-2025-11-12} with temperature 1.0 (the model default). The agent workspace directory structure is standardized across all runs, with source files organized to enable deterministic file navigation. For multi-app experiments, nested directory structures simulate realistic desktop organization (Desktop, Documents, Downloads). Each experiment is capped at 6 hours runtime with a theoretical limit of 25,000 tool calls.

\noindent\textbf{Task Load Categories.} We evaluate at four load levels via stratified sampling: 25\% ($\sim$12 tasks), 50\% ($\sim$23 tasks), 75\% ($\sim$35 tasks), and 100\% (all 46 tasks). For long-horizon evaluation, multiple tasks are concatenated into a single system prompt. At 100\% load, sustained execution spans multiple hours.

\subsection{Component Ablation Study}
\label{subsec:ablation}
To assess the contribution of individual architectural components under realistic operating conditions, we conduct an ablation study in the full MHTE setting. Rather than isolating tasks into separate prompts, we evaluate each configuration under 100\% task load, where all OSWorld Office tasks are concatenated into a single persistent execution session. This design measures how each component contributes to sustained performance under long-horizon, multi-task execution, which is the primary regime of interest in this work.

\noindent\textbf{Configurations.} We evaluate the cumulative addition of \corpgen components, starting from \textbf{UFO2}, a standalone computer-use agent. We then introduce a \textbf{Cognitive Model (orchestration layer)} that enables hierarchical planning, context management, and summarization via the \react reasoning loop. Next, we add \textbf{Cognitive Tools}, which provide explicit planning (e.g., generate plan, update plan), task tracking, and reflection capabilities. Finally, we incorporate \textbf{Experiential Learning (\corpgen)}, which introduces trajectory-based learning and agent-as-a-judge feedback, contingent on agent-as-judge integration.

Table~\ref{tab:component_ablation} reports task completion rates as components are added incrementally. Performance improves consistently across configurations, with the largest gain observed after introducing experiential learning, yielding the highest overall completion rate of approximately 15\%, driven primarily by improvements on Word and PowerPoint tasks.

\begin{table}[ht]
\caption{Component Ablation. Task completion (\%) at 100\% load across application-scoped sessions: Excel (11), Word (9), PPT (7), Multi-App (19). Total sums all 46 tasks.}
\label{tab:component_ablation}
\centering
\resizebox{\columnwidth}{!}{
\renewcommand{\arraystretch}{1.4}
\begin{tabular}{@{}lccccc@{}}
\toprule
 & \textbf{Excel} & \textbf{Word} & \textbf{PPT} & \textbf{Multi-App} & \textbf{Total} \\
\midrule
UFO2 & 0.0 & 0.0 & 14.3 & 5.3 & 4.3 \\
\quad + Cognitive Model & 0.0 & 33.3 & 0.0 & 5.3 & 8.7 \\
\quad + Cognitive Tools & 9.1 & 22.2 & 0.0 & 5.3 & 8.7 \\
\addlinespace[3pt]
\quad + Exp.\ Learning (\corpgen) & \textbf{9.1} {\small\color{teal}(+9.1)} & \textbf{33.3} {\small\color{teal}(+33.3)} & \textbf{28.6} {\small\color{teal}(+14.3)} & \textbf{5.3} {\small\color{gray}(--)} & \textbf{15.2} {\small\color{teal}(+10.9)} \\
\bottomrule
\end{tabular}
}
\end{table}

\noindent\textbf{Observation on Cognitive Model vs. Cognitive Tools.} Across short-duration, high-load experiments, we observe limited incremental gains from adding Cognitive Tools on top of the Cognitive Model. Both configurations achieve similar performance, suggesting that under these conditions the benefits of structured planning tools and model-level hierarchical reasoning may substantially overlap. In this regime, clear improvements emerge only after incorporating experiential learning.

\subsection{Task Load Scalability}
\label{subsec:load_scalability}
The central experiment asks: Does performance degrade as task load increases from 25\% to 100\%? We evaluate multiple computer using agent (CUA) frameworks (\textit{computer‑use‑preview-2025-03-11} and \textit{hierarchical computer‑use‑preview-2025-03-11}, which represent non‑orchestrated and hierarchically structured variants of the same CUA baseline) to demonstrate that this degradation is a general problem. We then show that \corpgen mitigates it. We measure task completion as task load increases from 25\% to 100\% by concatenating OSWorld tasks into a single persistent execution session. Baseline CUAs are compared against \corpgen using the same underlying CUAs to isolate the effect of orchestration.

\begin{table}[ht]
\caption{Task Load Scalability (task completion \%). All applications concatenated into a single session per load level: 25\% (${\sim}$12 tasks), 50\% (${\sim}$23), 75\% (${\sim}$35), 100\% (46). CUP = computer-use-preview.}
\label{tab:load_scalability}
\centering
\resizebox{\columnwidth}{!}{
\renewcommand{\arraystretch}{1.4}
\begin{tabular}{@{}lcccc@{}}
\toprule
\textbf{Configuration} & \textbf{25\%} & \textbf{50\%} & \textbf{75\%} & \textbf{100\%} \\
\midrule
UFO2 & 8.3 & 8.7 & 4.3 & 4.3 \\
CUP & 16.7 & 13.0 & 11.4 & 8.7 \\
Hierarchical CUP & 25.0 & 17.4 & 12.9 & 14.1 \\
\addlinespace[3pt]
\corpgen (UFO2) & \textbf{8.3} {\small\color{gray}(--)} & \textbf{17.4} {\small\color{teal}(+8.7)} & \textbf{11.4} {\small\color{teal}(+7.1)} & \textbf{8.7} {\small\color{teal}(+4.4)} \\
\corpgen (CUP) & \textbf{16.7} {\small\color{gray}(--)} & \textbf{13.0} {\small\color{gray}(--)} & \textbf{14.3} {\small\color{teal}(+2.9)} & \textbf{16.3} {\small\color{teal}(+7.6)} \\
\corpgen (H-CUP) & \textbf{25.0} {\small\color{gray}(--)} & \textbf{21.7} {\small\color{teal}(+4.3)} & \textbf{20.0} {\small\color{teal}(+7.1)} & \textbf{17.4} {\small\color{teal}(+3.3)} \\
\bottomrule
\end{tabular}
}
\end{table}

As shown in Table~\ref{tab:load_scalability}, baseline CUA performance declines as the number of concurrent tasks increases, while \corpgen mitigates degradation with higher completion rates at moderate and high task loads. These trends hold across multiple \corpgen settings (UFO2, computer-use-preview and Hierarchical computer-use-preview), indicating that the observed improvements stem from the orchestration layer rather than from any specific base agent.

\subsection{Application-Based Long-Horizon Evaluation}
\label{subsec:app_eval}
To complement the stratified task load evaluation, we evaluate long-horizon performance on application-based concatenated prompts. In this configuration, all tasks for a single application (e.g., all 11 Excel tasks or all 9 Word tasks) are concatenated into one prompt, testing sustained execution within a single application context.

\begin{table}[ht]
\caption{Application-based Long-Horizon Evaluation (Task Completion \%). All tasks for each application concatenated into a single prompt.}
\label{tab:per_app_longhorizon}
\centering
\scriptsize
\begin{tabular}{@{}lcccc@{}}
\toprule
& \textbf{Excel} & \textbf{Word} & \textbf{PPT} & \textbf{Multi-App} \\
& (11 tasks) & (9 tasks) & (7 tasks) & (19 tasks) \\
\midrule
UFO2 & 9.1\% & 22.2\% & 14.3\% & 5.3\% \\
\corpgen (UFO2) & \textbf{9.1\%} {\tiny\color{gray}(--)} & \textbf{22.2\%} {\tiny\color{gray}(--)} & \textbf{42.9\%} {\tiny\color{teal}(+28.6)} & \textbf{5.3\%} {\tiny\color{gray}(--)} \\
\bottomrule
\end{tabular}
\end{table}

This evaluation isolates the effect of task count within a consistent application context. Unlike the stratified evaluation (Table~\ref{tab:load_scalability}), which mixes applications, application-based evaluation reveals whether performance degradation stems from context switching between applications or from accumulated task complexity within a single application.

Relative to the baseline, Table~\ref{tab:per_app_longhorizon} shows that \corpgen improves completion rates in application-specific settings, with gains most evident in PowerPoint, while Excel and Word performance remains unchanged.

\section{Discussion}
\label{discussion}

\subsection{Validation of the MHTE Failure Modes}

Our results validate the four failure modes predicted in Section~\ref{problem-formulation}. The consistent degradation pattern across three independent CUA implementations (Table~\ref{tab:load_scalability}) demonstrates that the problem stems from fundamental architectural mismatches rather than implementation-specific issues. As task load increases from 25\% to 100\%, baseline CUAs show degradation across all architectures: UFO2 (8.3\%$\rightarrow$4.3\%), computer-use-preview (16.7\%$\rightarrow$8.7\%), and hierarchical (25.0\%$\rightarrow$14.1\%). This pattern directly reflects the predicted failure modes: context saturation forces lossy summarization, memory interference causes task conflation, dependency complexity overwhelms single-objective planners, and reprioritization overhead consumes reasoning capacity.

\corpgen's architecture addresses each failure mode through targeted mechanisms. Hierarchical planning maintains coherence across interleaved tasks by amortizing reprioritization decisions and explicitly tracking dependency graphs. Sub-agent isolation prevents memory interference by partitioning task-specific context. Tiered memory mitigates context saturation through selective retrieval and structured persistence. Adaptive summarization manages context growth while preserving critical information. The ablation study (Table~\ref{tab:component_ablation}) shows that these mechanisms work synergistically: the full architecture achieves 15.2\% completion versus 4.3\% for minimal baselines, with experiential learning providing the largest incremental gain.

\subsection{Qualitative Insights}

Failure analysis reveals that the predicted failure modes manifest as observable breakdowns: context overflow despite summarization, goal drift where actions diverge from strategic objectives, and cascading errors where early task failures propagate to dependent tasks. These effects only emerge under long-horizon evaluation with concurrent tasks, explaining why single-task benchmarks fail to expose them. Retry and skip policies prove critical for preventing infeasible tasks from stalling execution, while memory retrieval enables reuse of earlier context across task boundaries.

Experiential learning shows that routing feedback directly to the computer using agent is more effective than routing it through the hierarchical planner, and that a single hierarchical planning tool subsumes explicit daily planning in short-horizon settings, while longer deployments benefit from additional temporal structure. Sustained execution leads to emergent efficiency behaviors such as application reuse, operation batching, and cached context, reinforcing the importance of evaluating agents under realistic, persistent workloads rather than isolated tasks.

\subsection{Methodological Observation: Evaluation Approaches}

In preliminary experiments on evaluation methodology (separate from \corpgen), we observed that artifact-based judgment achieves substantially higher agreement with human labels (approximately 90\%) compared to trace-based LLM judgment (approximately 40\%). This suggests that standard evaluation methods relying solely on execution traces and screenshots may systematically underestimate long-horizon agent performance, as visual information alone proves inadequate for assessing task correctness. Effective judgment appears to require examining actual output artifacts (e.g., Excel files, documents) rather than observing actions alone. This observation motivates future work on evaluation methodologies for long-horizon agent assessment.

\subsection{Limitations}

Our framework inherits several limitations from current computer using agents (CUAs). Screenshot-based interaction provides broad applicability but remains slower and less reliable than structured API access, suggesting that hybrid approaches combining APIs with visual understanding are a more practical direction. Some CUA architectures achieve near-human performance~\cite{gonzalez-pumariega-2025-scaling} on the OSWorld benchmark, but they rely on a brute-force, bottom-up approach that is inefficient and not suited to the real-world setting we target with \corpgen. Our goal is to simulate a corporate environment where digital employees complete tasks efficiently and move on when they cannot be completed, mirroring how work proceeds in practice.

CUAs also struggle with context retention across task attempts and long workflows, leading to repeated failures when tasks are retried without memory of prior outcomes. Many errors further arise from incorrect assumptions about application state, such as acting before UI elements are ready or misinterpreting disabled controls, highlighting the need for stronger state verification.

Evaluation presents an additional challenge: judging task success requires access to rich contextual signals, including application state and produced artifacts, as AI judges tend to be overconfident when relying on limited traces. Beyond CUA limitations, our system incurs computational overhead from maintaining persistent digital employees and lacks ground-truth corporate behavioral data for evaluation. Finally, practical constraints such as vision API limits on the number of images per request required bounding execution traces during evaluation, which should be considered when designing long-horizon benchmarks.

\section{Conclusion}
\label{conclusion}

Existing long-horizon agent benchmarks evaluate sequential task persistence but fail to capture concurrent task management, the defining challenge of real-world organizational work. We introduced Multi-Horizon Task Environments (MHTEs) as a distinct problem class requiring agents to simultaneously manage dozens of interleaved long-horizon tasks with cross-task dependencies and dynamic reprioritization. Through empirical evaluation across three independent computer using agent (CUA) implementations, we demonstrated that single-task long-horizon agents experience catastrophic performance degradation in MHTEs (16.7\%$\rightarrow$8.7\% completion as load increases), validating four predicted failure modes: bounded context window saturation ($O(N)$ vs $O(1)$ context growth), cross-task memory interference (task conflation), dependency graph complexity (DAG vs linear chains), and reprioritization overhead ($O(N)$ vs $O(1)$ per-cycle decisions).

We presented \corpgen, an architecture-agnostic framework that addresses each failure mode through targeted mechanisms: hierarchical planning to manage dependency graphs and reduce reprioritization overhead, sub-agent isolation to prevent memory interference, tiered memory for cross-task state persistence, and adaptive summarization to control context growth. Evaluation demonstrates that \corpgen achieves up to 3.5$\times$ improvement over baselines (15.2\% vs 4.3\%), with consistent gains across three CUA backends, validating that improvements stem from architectural mechanisms rather than specific implementations. Ablation studies show that experiential learning provides the largest performance gains among architectural components.

These results establish MHTEs as a fundamental problem class distinct from single-task long-horizon reasoning and demonstrate that targeted architectural interventions can mitigate catastrophic degradation. The modular, CUA-agnostic design enables future work on organizational dynamics, multi-agent coordination, and extended deployments while supporting iterative improvement of individual components without system-wide redesign.

\makeatletter
\let\pra@section\section
\renewcommand{\section}{\@ifstar{\pra@starsection}{\pra@section}}
\newcommand{\pra@starsection}[1]{%
  \bigskip\noindent{\large\bfseries #1}\par\medskip}
\makeatother

\section*{Impact Statement}
This work advances research on long-horizon autonomous agents by introducing Multi-Horizon Task Environments (MHTEs) and a framework for simulating realistic organizational settings. The primary impact is enabling more rigorous study and evaluation of agents that must manage many concurrent, interleaved tasks over extended execution. \corpgen is intended as a research and simulation framework rather than a deployable system. It operates in controlled, synthetic environments and does not rely on real corporate data or model human workers. Potential risks associated with workplace automation or misinterpretation of agent capabilities are mitigated by this scope and by explicit artifact-based evaluation. In general, this work aims to improve the reliability, evaluation, and understanding of long-term agent behavior, supporting the safer and more transparent development of agentic systems.

\bibliographystyle{unsrt}
\bibliography{references}

\newpage
\appendix
\section{Evaluation Methodology}
\subsection{Judge Reliability and Meta-Evaluation}

To assess evaluation reliability, we conducted a small-scale meta-evaluation on 11 task executions (artifacts), sampled from the OSWorld Office benchmark. Ground-truth labels were provided by the authors based on task specifications and output artifacts. Agreement is measured as exact match between the judge’s success/failure decision and the human label. We compare an LLM-based judge operating on execution traces and screenshots against an agent-based judge that inspects task artifacts (e.g., generated files). The LLM judge uses the same base model as the execution agent. Artifact-based judging yields substantially higher agreement, reflecting reduced partial observability compared to trace-only evaluation

\end{document}